\documentclass[letterpaper, 10 pt, conference]{ieeeconf}  

\IEEEoverridecommandlockouts                              

\usepackage{graphicx}
\usepackage{xcolor}
\usepackage{subcaption}
\usepackage{colortbl}
\usepackage{amsmath}
\DeclareMathOperator*{\argmin}{arg\,min}

\usepackage{pifont}
\newcommand{\cmark}{\textcolor{green}{\ding{51}}}%
\newcommand{\xmark}{\textcolor{red}{\ding{55}}}%

\title{\LARGE \bf
Revisiting PatchMatch Multi-View Stereo for Urban 3D Reconstruction
}

\author{Marco Orsingher$^{1}$, Paolo Zani$^{2}$, Paolo Medici$^{2}$ and Massimo Bertozzi$^{3}$
\thanks{$^{1}$Marco Orsingher is with Department of Engineering and Architecture, University of Parma, Italy, and VisLab Srl, an Ambarella Inc company
        {\tt\small marco.orsingher@unipr.it}}
\thanks{$^{2}$Paolo Zani and Paolo Medici are with VisLab Srl, an Ambarella Inc company}
\thanks{$^{3}$Massimo Bertozzi is with Department of Engineering and Architecture, University of Parma, Italy}
}

\begin{document}

\maketitle
\thispagestyle{empty}
\pagestyle{empty}

\begin{abstract}

In this paper, a complete pipeline for image-based 3D reconstruction of urban scenarios is proposed, based on PatchMatch Multi-View Stereo (MVS). Input images are firstly fed into an off-the-shelf visual SLAM system to extract camera poses and sparse keypoints, which are used to initialize PatchMatch optimization. Then, pixelwise depths and normals are iteratively computed in a multi-scale framework with a novel depth-normal consistency loss term and a global refinement algorithm to balance the inherently local nature of PatchMatch. Finally, a large-scale point cloud is generated by back-projecting multi-view consistent estimates in 3D. The proposed approach is carefully evaluated against both classical MVS algorithms and monocular depth networks on the KITTI dataset, showing state of the art performances.


\end{abstract}

\section{INTRODUCTION}

Accurate and complete 3D understanding of the surrounding environment is one of the main challenges in autonomous driving. The classical sensor fusion approach requires a complex sensor suite that is both expensive and challenging to calibrate. Therefore, in recent years, there is a growing interest in achieving this goal with simple and low cost sensors, such as cameras. Image-based 3D reconstruction of large-scale scenes is usually tackled in literature from two different perspectives: end-to-end deep learning methods and the classical two stages pipeline, composed by Structure From Motion (SFM) and Multi-View Stereo (MVS).

On one hand, monocular depth networks (\cite{monodepth2, packnet, dorn, bts}) are the most popular approach for urban scenarios, as they can be trained to predict depth using appearance and relative geometry as the only source of supervision. While dealing effectively with problems like dynamic objects, these methods suffer from severe generalization issues to arbitrary camera configurations, as they learn the image-to-depth mapping with the camera parameters used for training.  

On the other hand, classical Multi-View Stereo algorithms (\cite{acmm, acmp, deepcmvs}), mostly based on the PatchMatch framework (\cite{patchmatch}, \cite{pmstereo}), still dominate the leaderboards (\cite{eth3d, tanks}) when reconstructing complex and arbitrary outdoor scenes. These approaches rely on multi-view geometry equations, which naturally generalize to novel situations. However, matching pixels across multiple images is prone to errors in textureless areas and non-Lambertian surfaces. Moreover, the local optimization procedure of PatchMatch fails to propagate good solutions in these regions, thus producing artifacts and noise, despite recent algorithmic efforts (\cite{acmm, acmp, tapamvs}).

In this paper, common failure modes of PatchMatch MVS are tackled in the context of urban scenarios, leading to a reliable framework for outdoor monocular 3D reconstruction. Firstly, sparse keypoints from visual SLAM are used to initialize the 3D geometry. Then, a novel depth-normal consistency loss function is designed to enforce more accurate solutions. Finally, the algorithm iterates between local PatchMatch optimization and a global refinement step in a multi-scale fashion. Pixelwise depths and normals are computed for each frame and a dense point cloud of the scene is obtained from multi-view consistent estimates.

The proposed approach has been evaluated on classical urban datasets and compared with monocular depth networks, since little work (\cite{monorec, semantically}) has been done in literature to benchmark MVS for urban data. Moreover, the effect of each contribution on the resulting 3D model is carefully analyzed in ablation studies.

The rest of the paper is organized as follows. Related works and novel contributions are presented in Sec.~\ref{sec:rel}. Then, the proposed approach is detailed in Sec.~\ref{sec:method}, where each subsection is devoted to a revisited component of the algorithm. Finally, experimental results are shown in Sec.~\ref{sec:res} and conclusions are drawn in Sec.~\ref{sec:conc}.

\begin{figure*}
    \centering
    \includegraphics[width=\textwidth]{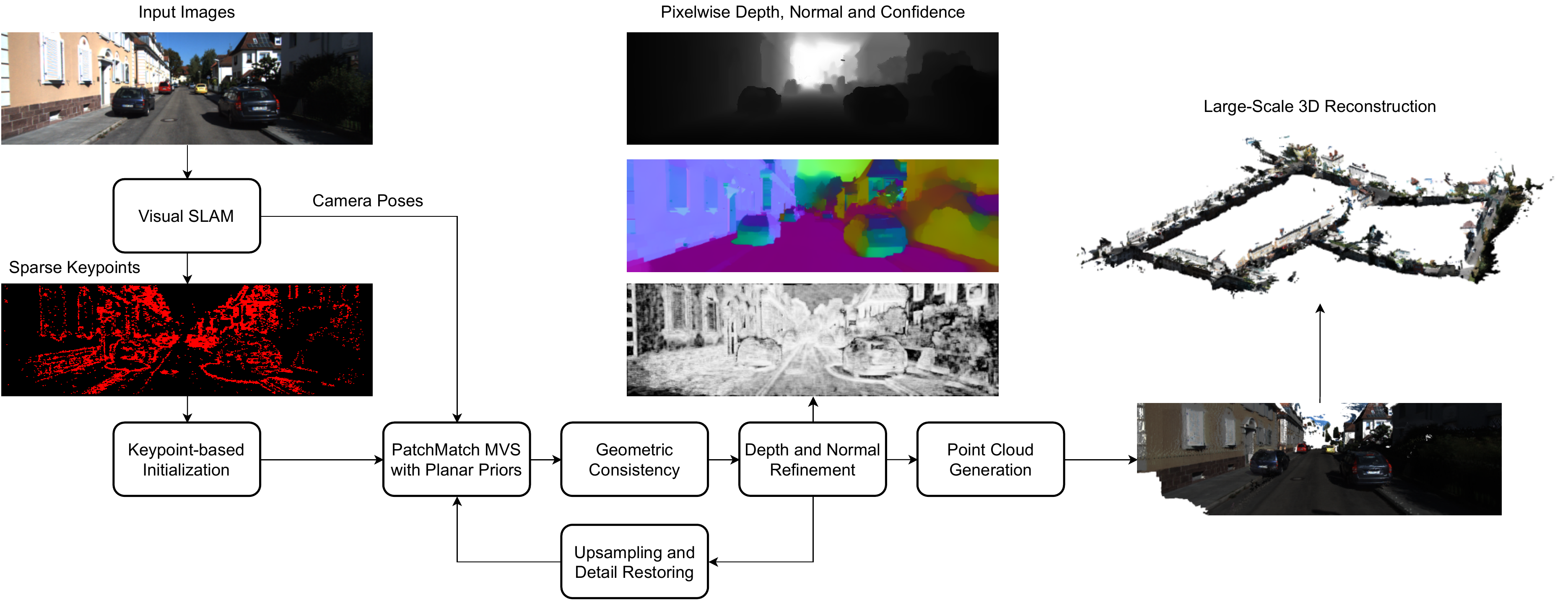}
    \caption{Overview of the proposed approach for large-scale 3D reconstruction of urban environments from images.}
    \label{fig:overview}
\end{figure*}

\section{RELATED WORK}
\label{sec:rel}

\subsection{Monocular Depth Networks}

Initial works in learning-based monocular depth estimation designed convolutional architectures for regressing depth from images with ground truth supervision (\cite{eigen, dorn}). Despite showing promising results, acquiring ground truth data for depth requires additional expensive sensors (e.g.\ LiDAR) and the supervision is rather sparse and noisy. 

In order to avoid this issue, Zhou \textit{et al.}~\cite{zhou} pioneered a fully self-supervised approach to jointly learn depth and pose by using view synthesis as a proxy task. Recent improvements in network architectures and loss functions (\cite{monodepth2, packnet, midas}) have shown performances almost on par with supervised methods. Another hybrid line of research (\cite{sparse_1, sparse_2}) explores the possibility of using sparse LiDAR data as a weak supervision for scale-aware monocular networks.

The main advantage of these methods is that they learn a direct image-to-depth mapping, which can be directly deployed for real-time navigation of autonomous vehicles. However, neural networks are usually limited in the input resolution and they require fine-tuning on unseen data, as they do not generalize zero-shot to different camera configurations. Furthermore, pixelwise normals are typically not handled, despite few notable exceptions (\cite{lego, normal_1}). This is a critical issue when the downstream task is 3D reconstruction, since the joint estimation of depth and normals improves the geometric consistency of the resulting model.

\subsection{PatchMatch Multi-View Stereo}

PatchMatch Multi-View Stereo algorithms belong to the depth-map-based class of MVS methods~\cite{survey}, as they estimate a dense depth and normal map for each input view, by exploiting the core idea of PatchMatch~\cite{patchmatch} to sample and propagate good hypotheses. 

Starting from the seminal work on PatchMatch Stereo~\cite{pmstereo}, early approaches in the field focused on improving the core building blocks of the algorithm. Sequential propagation was proposed in~\cite{colmap}, as well as pixelwise view selection with a probabilistic graphical model. For dealing with efficiency issues, Galliani \textit{et al.}~\cite{gipuma} introduced the red-black propagation checkerboard scheme, which has been subsequently improved by~\cite{acmh} with adaptive sampling. This allows for massive parallelization on GPUs as half of the pixels are simultaneously updated.

Some approaches have been proposed to discriminate good solutions in textureless areas, where photometric consistency is unreliable: multi-scale estimation~\cite{acmm}, planar prior models~\cite{acmp}, multi-view geometric consistency~(\cite{acmm, colmap}) and superpixels extraction~\cite{tapamvs}. This work combines planar prior models with an improved geometric consistency check into a unified multi-scale framework, thus inheriting the advantages of such modules.

A recent work~\cite{semantically} designed a semantically guided PatchMatch MVS method for 3D reconstruction of urban scenarios. However, it strongly relies on the output of a semantic segmentation network~\cite{deepvlab3} during each stage of the pipeline, while the presented approach only requires raw images. 

\subsection{Contributions}

From an algorithmic point of view, the contribution of the presented method to existing literature on PatchMatch MVS is threefold: (i) the hypotheses initialization step is augmented with sparse keypoints information (Sec.~\ref{sec:init}); (ii) a novel geometric loss function is introduced to ensure consistency between depths and normals (Sec.~\ref{sec:cons}); (iii) a multi-scale interaction between the local optimization of PatchMatch and a global refinement algorithm is proposed to regularize the solution (Sec.~\ref{sec:ref}).

Furthermore, a detailed quantitative and qualitative evaluation of PatchMatch MVS on street-level data, as well as a comparison with monocular depth networks, are provided (Sec.~\ref{sec:res}). To the authors' knowledge, this is the first comparative analysis of these two approaches for urban scenarios.

\section{PROPOSED APPROACH}
\label{sec:method}

The framework takes a set of $N$ sequential images as input, typically as the frames of a monocular video. Firstly, an off-the-shelf visual SLAM system~\cite{dvso} is used to extract camera poses and sparse keypoints; those serve as initialization for PatchMatch MVS\@. Then, pixelwise estimates of depth, normal and cost are computed for each image. Finally, a point cloud is generated by back-projecting all the pixels with consistent 3D hypotheses among multiple views. Without loss of generality, the reference image (i.e.\ the current frame) and the source images (i.e.\ neighboring frames) are denoted as $I_0$ and $\{I_i\}_{i=1}^{K}$, respectively (K = 4). An overview of the proposed approach is shown in Fig.~\ref{fig:overview}.

\subsection{PatchMatch MVS with Planar Priors}
\label{sec:pm}

The presented method is built upon the PatchMatch MVS algorithm with planar priors introduced in~\cite{acmp} and referred in the following as \emph{baseline}. In this section, a concise description of the baseline approach is provided for the sake of completeness and the reader is referred to~\cite{acmp} for more details. Starting from a random initial solution, the algorithm iterates over three steps until convergence: propagation, evaluation and perturbation. 

During the propagation phase, the red-black checkerboard scheme~\cite{gipuma} is used to gather a set of $K$ candidate hypotheses $\mathcal{S} = \{(d_j, \mathbf{n}_j)\}_{j=1}^K$ for each pixel $\mathbf{p}$ from its neighbors with adaptive sampling \cite{acmh}. The goal is to find the depth $d$ and normal vector $\mathbf{n}$ that minimize a pixelwise photometric matching cost:

\begin{equation}
    (d^*,\mathbf{n}^*) = \argmin_{(d,\mathbf{n}) \in \mathcal{S}} \mathcal{L}_{photo}(d,\mathbf{n})
\end{equation}

Intuitively, each hypothesis represents the local tangent plane to the scene surface at the 3D point corresponding to the pixel $\mathbf{p}$. To this end, the current pixel is warped with planar homography onto each source view $i$ for each candidate hypothesis $j$ and the cost $c_{ij}$ is computed with normalized cross-correlation~\cite{colmap}. Then, view selection and multi-view cost aggregation~\cite{acmh} are performed to produce a single cost $\Bar{c}_j = \mathcal{L}_{photo}(d,\mathbf{n})$ per hypothesis. The solution is updated with the plane minimizing such cost. 

Finally, in the perturbation step a small set of new hypotheses, built by combining current, perturbed and random depth and normals, is tested in order to refine the estimates and escape bad local minima.

After $N_{photo}$ iterations, sparse but reliable correspondences are computed and triangulated to generate planar priors for each pixel. In particular, a Delaunay triangulation is built upon pixels with cost lower than 0.1 and a plane for each triangle is computed using its three vertices projected in 3D. The same three steps above are then repeated for $N_{planar}$ iterations and the matching cost is extended to include such priors as $\mathcal{L}_{photo}(d,\mathbf{n}) + \mathcal{L}_{planar}(d,\mathbf{n})$. The definition of the planar priors cost follows the original formulation in~\cite{acmp}.

The proposed framework, differently from~\cite{acmp}, casts this procedure into a coarse-to-fine multi-scale scheme to further improve the solution in textureless areas. The joint bilateral upsampler~\cite{jbu} and the detail restorer proposed in~\cite{acmm} are adopted on three hierarchy levels.

\subsection{Keypoint-based Initialization}
\label{sec:init}

The usual practice in existing literature (\cite{gipuma, colmap, acmh, acmm, acmp}) is to perform PatchMatch optimization starting from a random initial solution. More specifically, depth candidates are randomly sampled for each pixel within a pre-defined range $[d_{min}, d_{max}]$, while random unit normals can be efficiently generated by following~\cite{gipuma}. 

However, it is common to have a depth guess for at least a sparse subset of pixels. This initial solution might originate from multiple sources and the \textbf{first contribution} is to look deeper into how such priors can be integrated in the PatchMatch pipeline, under different assumptions. In the proposed approach, the visual SLAM pipeline outputs a sparse point cloud, which is then reprojected onto each image and used to bootstrap the optimization.

Additionally, in urban scenarios the sensor suite of a self-driving car typically includes range sensors or stereo cameras that can provide an approximate depth estimate. In a more general 3D reconstruction pipeline, well-triangulated 2D features and corresponding 3D keypoints are a by-product of most SFM algorithms~\cite{colmap2}. The PatchMatch framework is flexible enough to support any kind of initial solution, without requiring to design complex neural architectures to extract representations from sparse input data~\cite{boosting}.

\begin{figure}
    \centering
    \includegraphics[width=\linewidth]{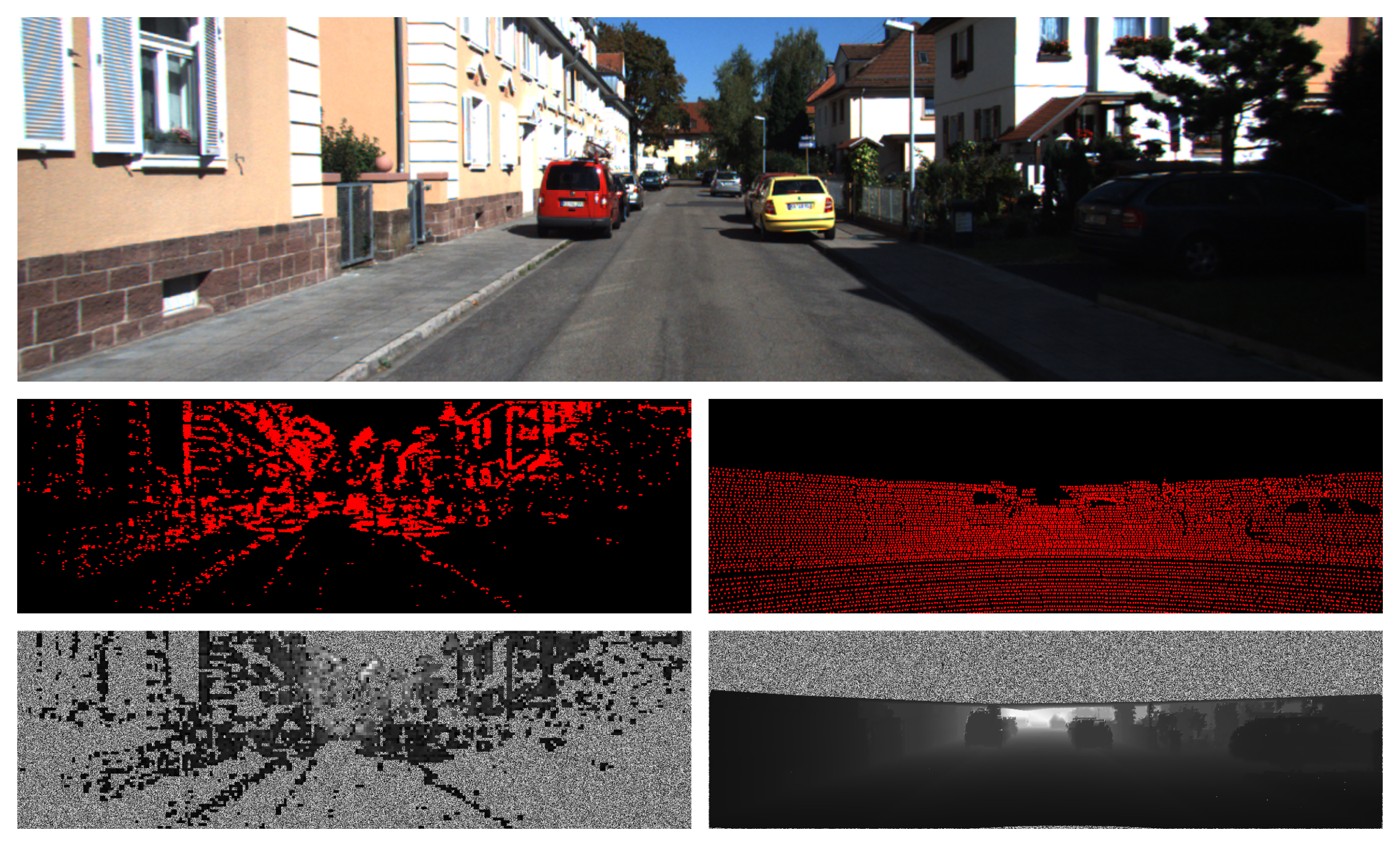} 
    \caption{Examples of proposed initialization with densified visual SLAM keypoints (left) and triangulated sparse LiDAR points (right) for the input image (top). Raw measurements and the corresponding initial depth are shown in the center and bottom row, respectively.}
    \label{fig:init}
\end{figure}

Classical MVS datasets (\cite{tanks, eth3d}) contain sequences of cameras with wide triangulation angles and mostly textured scenes, which produce a fairly regular distribution of keypoints in the image. In this case, previous works (\cite{marmvs, semantically}) suggest to triangulate the reprojected depth values and compute plane parameters for each triangle. When available, this approach can be applied in urban scenarios to sparse LiDAR points, as shown in Fig.~\ref{fig:init} (right).

On the other hand, keypoints from visual SLAM suffer from very low triangulation angles and they are usually concentrated in specific areas, while being almost absent in textureless regions, such as roads. Moreover, noisy and erroneous matches are much more frequent in real-world monocular scenarios, thus making the triangulation approach unstable and unreliable. Therefore, in this case the initial solution is simply densified, meaning that each depth value is propagated in a local support region of $5 \times 5$ pixels. The result is shown in Fig.~\ref{fig:init} (left).

In both cases, hypotheses are perturbed with random Gaussian noise to promote diversification and pixels without an associated triangle or too far from any keypoint are still randomly initialized. Since the proposed pipeline is agnostic to any kind of initial solution that can be provided, a detailed analysis on how the quality and the number of initial depth values influence the final estimates is presented in Sec.~\ref{sec:abl}.

\subsection{Improved Geometric Consistency}
\label{sec:cons}

Photometric consistency is known to fail in textureless areas, while the planar prior term is unreliable when erroneous sparse correspondences are established. To this end, some works enforce multi-view geometric consistency as the forward-backward reprojection error (\cite{colmap, acmm}). 

A pixel $\mathbf{p}_0$ in the reference image is first projected in 3D according to its current depth estimate and observed by a source image as the pixel $\mathbf{p}_i$. Then, this pixel is re-projected in 3D with the source depth hypothesis and observed again in the reference image to obtain the pixel $\hat{\mathbf{p}}_0$. The reprojection error $\mathcal{L}_{rep}(d,\mathbf{n})$ is given by $||\mathbf{p}_0 - \hat{\mathbf{p}}_0||$, clamped by a robust threshold $\tau$ against occlusions.

However, this formulation does not fully exploit the local planar geometry of the scene, since consistency is enforced for depth estimates only and not for normal vectors. Therefore, inspired by recent literature in monocular depth estimation (\cite{cons1, cons2}), the \textbf{second contribution} of this work is an additional loss term to drive the PatchMatch optimization procedure towards more consistent solutions. 

Specifically, for a given pixel $\mathbf{p}$, a unit normal vector $\mathbf{n}_\mathbf{p}$ is consistent with its corresponding depth value $d_\mathbf{p}$ if it is locally perpendicular to the surface at the corresponding 3D point $\mathbf{X_p} =\mathbf{X}(\mathbf{p}, d_\mathbf{p})$. Let $\Omega$ be a local support region around the pixel. A consistent planar solution must minimize the following cost:

\begin{equation}
    \mathcal{L}_{cons}(d,\mathbf{n}) = \sum_{\mathbf{q} \in \Omega} w(\mathbf{p},\mathbf{q}) \cdot \mathbf{n}_\mathbf{p}^\top (\mathbf{X_q} - \mathbf{X_p})
\end{equation}

where $w(\mathbf{p}, \mathbf{q}) = e^{-||I(\mathbf{q}) - I(\mathbf{p})||}$ downweights pixels with different color values, which are less likely to belong to the same surface. The depth-normal consistency is combined with the forward-backward reprojection error to obtain the final geometric consistency cost for each source image:

\begin{equation}
    \mathcal{L}_{geom}(d,\mathbf{n}) = \lambda_{rep}\mathcal{L}_{rep}(d,\mathbf{n}) + \lambda_{cons}\mathcal{L}_{cons}(d,\mathbf{n})
\end{equation}

After the photometric and planar iterations, PatchMatch optimization is repeated again $N_{geom}$ times to include this cost, weighted and aggregated by view selection priors~\cite{acmh}.


\begin{figure}
    \centering
    \includegraphics[width=\linewidth]{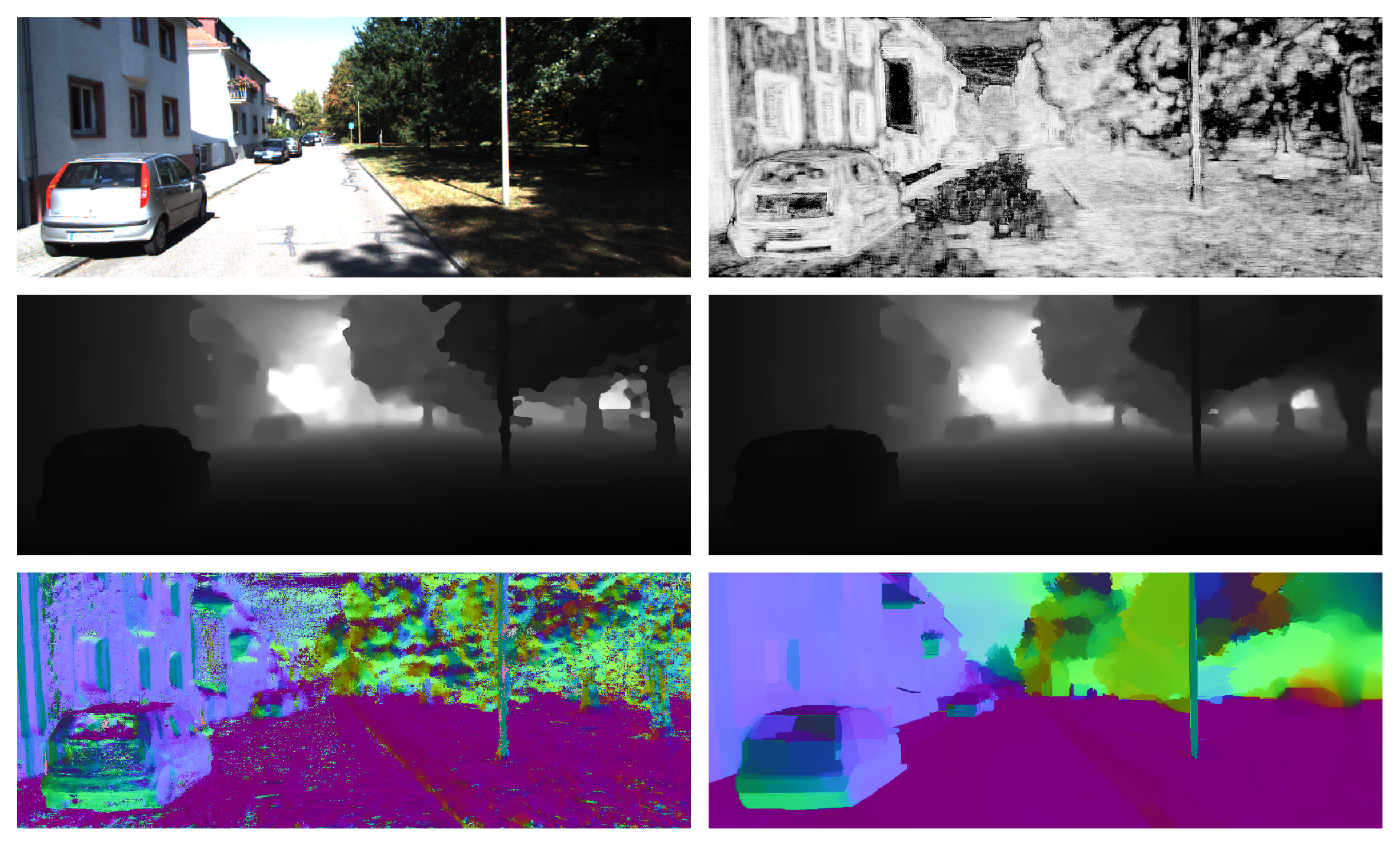}
    \caption{Comparison of raw (left) and refined (right) depths (central row) and normals (bottom row) computed for the input image with associated confidence (top row).}
    \label{fig:refine}
\end{figure}

\subsection{Depth and Normal Refinement}
\label{sec:ref}

Despite the efforts in ensuring photometric, planar and geometric consistency, the output depth and normal maps can still contain errors that can harm both the accuracy and completeness of the resulting 3D models. The baseline~\cite{acmp} and related works (\cite{acmh, acmm}) perform a simple median filtering to remove outliers after each PatchMatch iteration. 

However, the root cause of most errors is the inherently local nature of the PatchMatch algorithm, since the global context is never considered during optimization. To this end, the \textbf{third contribution} of this paper is to leverage the confidence-based global optimization algorithm presented in~\cite{refine} to refine the depth and normal estimates at each scale. Confidence maps are generated by clipping the cost of a given hypothesis between 0 (unreliable) and 1 (reliable). 

The refinement procedure optimizes both a unary term that penalizes deviations from reliable input values and a binary term that promotes globally smooth solutions. 
The reader is referred to~\cite{refine} for more details. 
An example of raw and refined outputs can be found in Fig. \ref{fig:refine}.

Note that while a previous work~\cite{deepcmvs} adopted the same refinement approach to MVS output, the key improvements of the presented method are (i) a much simpler confidence metric and (ii) the use of the algorithm \textit{at each scale} of the pyramid, rather than a single time at the end. This means that the multi-view consistency of refined solutions is further enforced at finer scales, thus producing more complete and accurate 3D models.

\begin{table*}
\caption{Quantitative results on the KITTI dataset. Best and second results are marked \textbf{bold} and \underline{underlined}, respectively. Yellow columns are error metrics where lower is better, orange columns are accuracy metrics where higher is better~\cite{eigen}.}
\label{tab:quant}
\small
\begin{center}
\begin{tabular}{|c|c|c|c|c|c|c|c|c|c|}
\hline
Method & Dataset & Resolution & \cellcolor{yellow}Abs Rel & \cellcolor{yellow}Sq Rel & \cellcolor{yellow}RMSE & \cellcolor{yellow}RMSE$_{log}$ & \cellcolor{orange} $\delta < 1.25$ & \cellcolor{orange} $\delta < 1.25^2$ & \cellcolor{orange} $\delta < 1.25^3$ \\
\hline\hline
MonoDepth2 \cite{monodepth2} & Eigen & 1024 $\times$ 320 & 0.115 & 0.882 & 4.701 & 0.190 & 0.879 & 0.961 & 0.982 \\
PackNet \cite{packnet} & Eigen & 1280 $\times$ 384 & 0.107 & 0.882 & 4.538 & 0.186 & 0.889 & 0.962 & 0.981 \\
ManyDepth \cite{manydepth} & Eigen & 1024 $\times$ 320 & 0.087 & 0.685 & 4.142 & 0.167 & 0.920 & 0.968 & 0.983 \\
DORN \cite{dorn} & Eigen & 513 $\times$ 385 & 0.077 & 0.290 & 2.723 & 0.113 & 0.949 & 0.988 & 0.996 \\
BTS \cite{bts} & Eigen & 704 $\times$ 352 & 0.059 & 0.245 & 2.756 & 0.096 & 0.956 & 0.993 & 0.998 \\
MiDAS \cite{midas} & Eigen & Full & 0.062 & - & 2.573 & 0.092 & 0.959 & 0.995 & \textbf{0.999}  \\
\hline
Colmap \cite{colmap} & Odom & Full & 0.099 & 3.451 & 5.632 & 0.184 & 0.952 & 0.979 & 0.986 \\
ACMM \cite{acmm} & Odom & Full & 0.042 & 0.498 & 2.871 & 0.166 & 0.982 & 0.989 & 0.992  \\
ACMP \cite{acmp} & Odom & Full & \underline{0.034} & 0.381 & \underline{1.930} & 0.152 & \underline{0.987} & 0.992 & 0.994  \\
DeepMVS \cite{deepmvs} & Odom & 512 $\times$ 256 & 0.088 & 0.644 & 3.191 & 0.146 & 0.914 & 0.955 & 0.982 \\
DeepTAM \cite{deeptam} & Odom & 512 $\times$ 256 & 0.053 & 0.351 & 2.480 & 0.089 & 0.971 & 0.990 & 0.995  \\
MonoRec \cite{monorec} & Odom & 512 $\times$ 256 & 0.050 & \underline{0.295} & 2.266 & \underline{0.082} & 0.973 & 0.991 & 0.996  \\
\hline
\textbf{Ours} & Odom & Full & \textbf{0.025} & \textbf{0.201} & \textbf{1.459} & \textbf{0.069} & \textbf{0.992} & \textbf{0.996} & \underline{0.998}\\
\hline
\end{tabular}
\end{center}
\end{table*}

\subsection{Point Cloud Generation}
\label{sec:pcl}

Given the depths and normals at the finest scale, a point cloud can be computed by fusing them into a consistent 3D model (\cite{gipuma, acmm}). Each input image is set as reference in turn and each pixel is projected to a 3D point in the world frame. Then, these points are observed by all the source images. 

A 3D point is defined as consistent and added to the point cloud if the following three conditions are met for at least $N_{min}$ views: (i) the forward-backward reprojection error is lower than $\gamma$, (ii) the relative depth difference is lower than $\epsilon$ and (iii) the angle between corresponding normals is lower than $\theta$. In that case, a unified 3D point with normal is computed by averaging all the consistent hypotheses.

\section{EXPERIMENTS}
\label{sec:res}


\subsection{Implementation Details}
\label{sec:impl}

The whole framework has been implemented in C++/CUDA and the code runs on a single NVIDIA GTX 1080Ti GPU. The following parameters have been used during experiments: $N_{photo} = N_{planar} = 3$,~$N_{geom} = 2$, $\lambda_{rep} = \lambda_{cons} = 0.1$, $\tau = 2$. For point cloud generation, consistency thresholds are set to~$N_{min} = 2$, $\gamma = 2$, $\epsilon = 0.01$ and $\theta = 10$°. 


\subsection{Quantitative Evaluation}
\label{sec:quant}

For quantitative evaluation on the KITTI dataset, the usual choice is the Eigen test split~\cite{eigen}. However, MVS methods require temporally adjacent frames with estimated poses. Following~\cite{monorec}, the presented algorithm has been evaluated on the intersection between the KITTI odometry benchmark and the Eigen test split (referred to as Odom in Tab.~\ref{tab:quant}). This procedure leads to 8634 images for testing. Moreover, camera poses and sparse keypoints for initialization are computed with the visual SLAM system DVSO~\cite{dvso}, in order to enable 3D reconstruction from images alone. All the results have been obtained at full resolution and clamped at 80~m for computing metrics. Consistently with recent literature, the improved ground truth depth maps from~\cite{sparsity} and the metrics from~\cite{eigen} are used for evaluation.

The first class of methods chosen for comparison is monocular depth networks. In order to cover a broad range of methods, both self-supervised (\cite{packnet, monodepth2, manydepth}) and fully supervised networks (\cite{dorn, bts}) have been selected, as well as the recent vision transformer-based approach~\cite{midas}. In addition, both classical (\cite{colmap, acmm, acmp}) and learning-based (\cite{monorec, deepmvs, deeptam}) MVS algorithms have been considered.

Tab.~\ref{tab:quant} shows that the proposed method outperforms state of the art by a notable margin. It can be seen that the performance gap with respect to monocular networks is shared by most MVS approaches, due to their ability of exploiting information about camera poses and adjacent frames, rather than directly trying to map images to depth. Moreover, the contributions presented in this work  allow to further boost performances with respect to other MVS algorithms as well.

\subsection{Ablation Studies}
\label{sec:abl}

In order to quantify more precisely the effect of each component on the final result, ablation studies are presented in Tab.~\ref{tab:abl}, showing that all the proposed contributions consistently improve the depth estimation metrics with respect to the baseline~\cite{acmp}. Specifically, the multi-scale estimation framework and the global refinement algorithm are the two components with the highest impact. This can be seen also with the comparison between the baseline and another work~\cite{deepcmvs} that exploits the same refinement step to MVS output. Keypoint-based initialization and the novel depth-normal consistency term allow to further boost the performances of the proposed method.

\begin{figure}
    \centering
    \includegraphics[width=\linewidth]{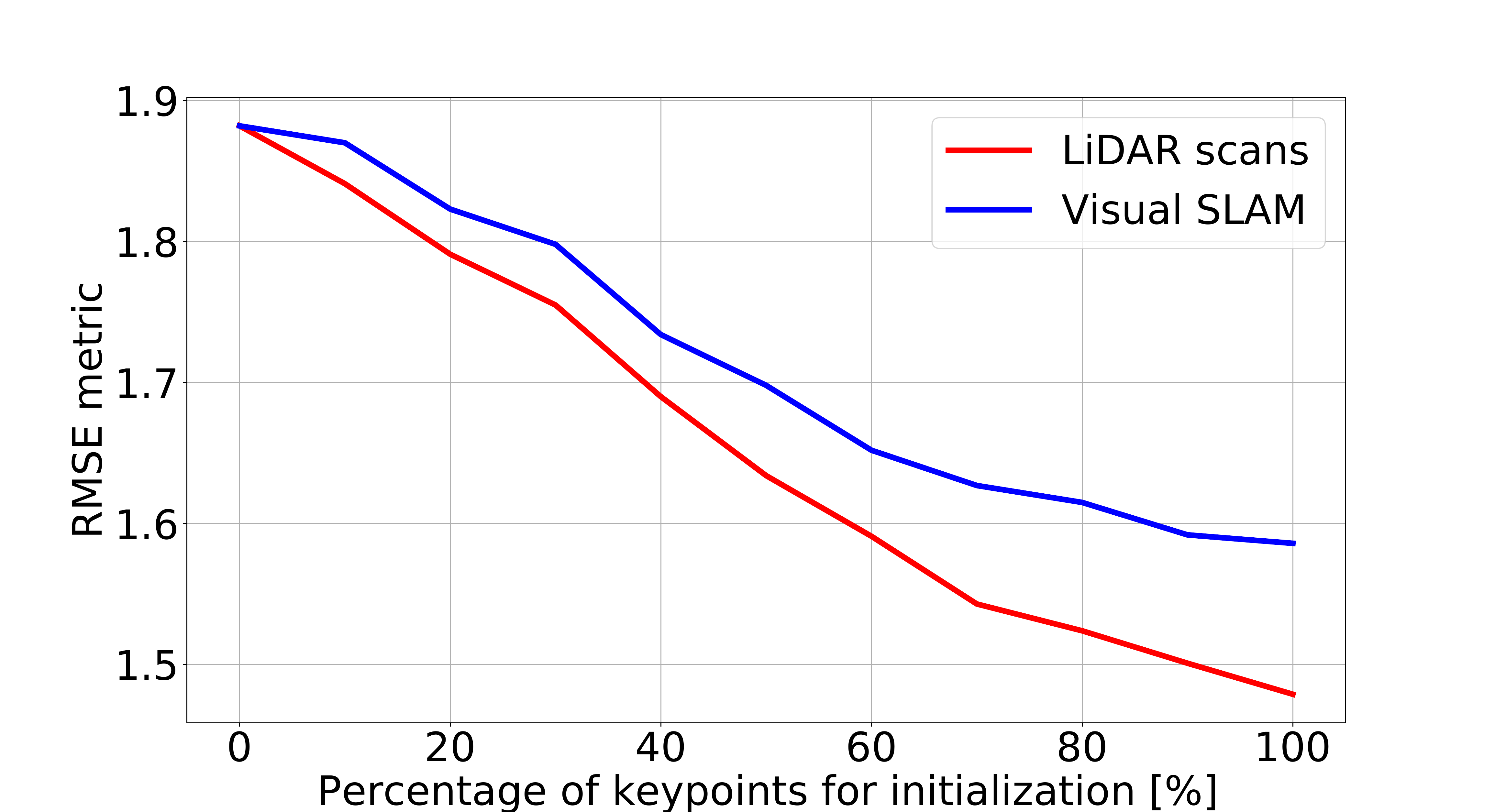} 
    \caption{RMSE metric on the KITTI dataset as a function of the number of input depth values for both LiDAR data and visual SLAM keypoints.}
    \label{fig:abl}
\end{figure}

Furthermore, the influence of the quantity and the quality of initial depth values on the final results have been deeper investigated. Fig.~\ref{fig:abl} shows the RMSE metric as a function of the number of points for both LiDAR and visual SLAM input data. Intuitively, having more keypoints almost always improve the solution. It can also be seen that LiDAR points are generally better than SLAM keypoints. This is due to the fact that LiDAR provides regular depth estimates in textureless regions where SLAM typically fails to extract keypoints and these are the same areas in which MVS struggles the most. Therefore, additional input from another sensing modality is beneficial. 

\begin{table*}
\caption{Ablation studies on the KITTI dataset. Best and second results are marked \textbf{bold} and \underline{underlined}, respectively. Yellow columns are error metrics where lower is better, orange columns are accuracy metrics where higher is better~\cite{eigen}. Legend: MS - Multi-Scale, KP - KeyPoint-based initialization, GC - our Geometric Consistency, Ref - global Refinement.}
\label{tab:abl}
\small
\begin{center}
\begin{tabular}{|c|c|c|c|c|c|c|c|c|c|c|c|}
\hline
Method & MS & KP & GC & Ref & \cellcolor{yellow}Abs Rel & \cellcolor{yellow}Sq Rel & \cellcolor{yellow}RMSE & \cellcolor{yellow}RMSE$_{log}$ & \cellcolor{orange} $\delta < 1.25$ & \cellcolor{orange} $\delta < 1.25^2$ & \cellcolor{orange} $\delta < 1.25^3$ \\
\hline\hline
Baseline \cite{acmp} & \xmark & \xmark & \xmark & \xmark & 0.034 & 0.381 & 1.930 & 0.152 & 0.987 & 0.992 & 0.994 \\
Baseline (ref) \cite{deepcmvs} & \cmark & \xmark & \xmark & \cmark & \underline{0.031} & 0.284 & 1.716 & \underline{0.074} & \underline{0.991} & \underline{0.995} & \underline{0.997} \\
\hline
Ours & \cmark & \xmark & \xmark & \xmark & 0.033 & 0.321 & 1.882 & 0.102 & 0.988 & 0.994 & 0.996 \\
Ours & \cmark & \cmark & \xmark & \xmark & 0.032 & 0.305 & 1.603 & 0.081 & 0.990 & \underline{0.995} & 0.997 \\
Ours & \cmark & \cmark & \cmark & \xmark & 0.032 & \underline{0.273} &  \underline{1.586} & 0.076 & \underline{0.991} & \textbf{0.996} & \underline{0.997} \\
\hline
\textbf{Ours (full)} & \cmark & \cmark & \cmark & \cmark & \textbf{0.025} & \textbf{0.201} & \textbf{1.459} & \textbf{0.069} & \textbf{0.992} & \textbf{0.996} & \textbf{0.998}\\
\hline
\end{tabular}
\end{center}
\end{table*}

\subsection{Qualitative Results}
\label{sec:qual}

\begin{figure*}
    \begin{subfigure}{0.19\textwidth}
      \centering
      \includegraphics[width=\linewidth]{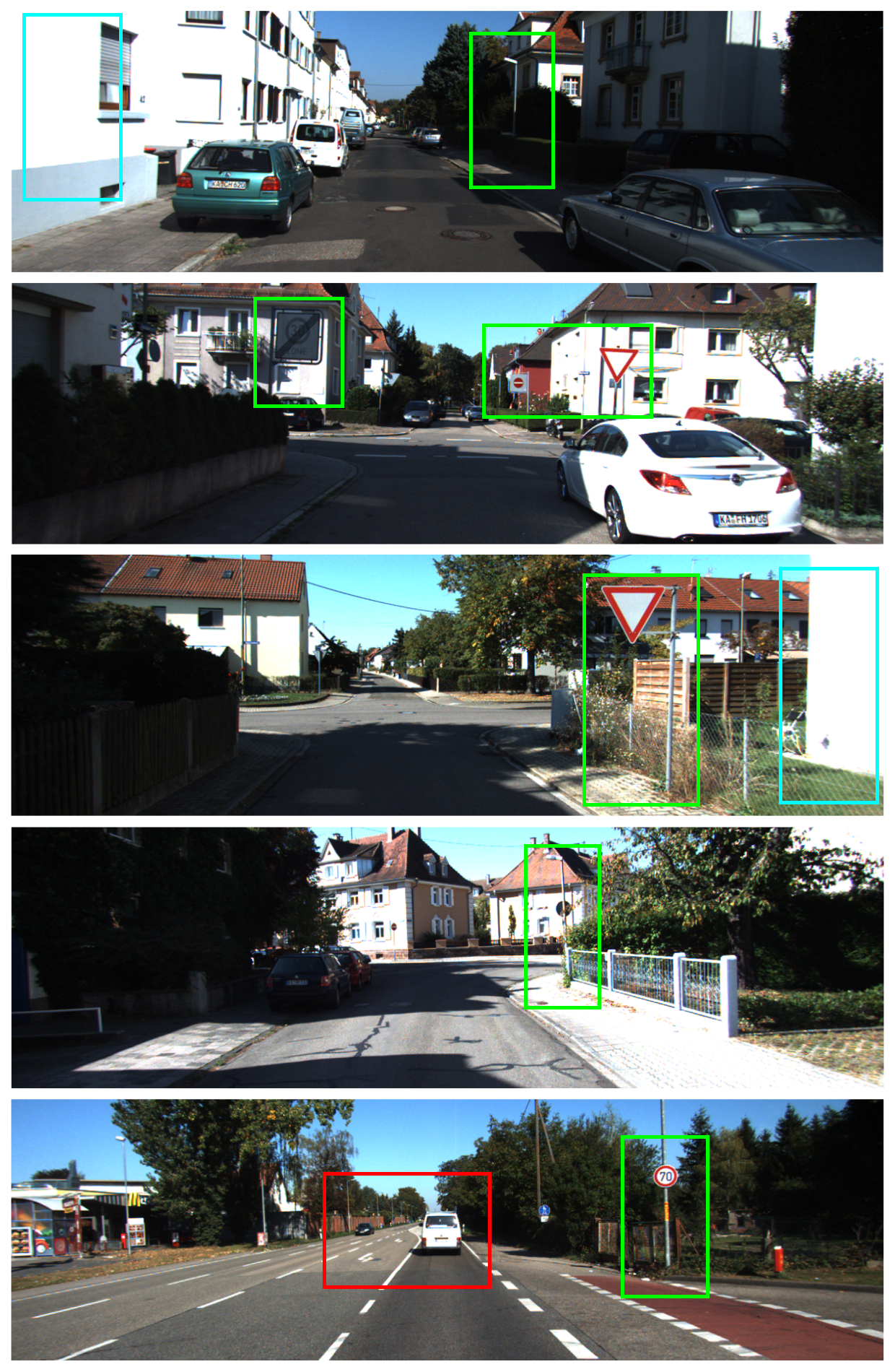}
      \caption{Input image}
    \end{subfigure}
    \begin{subfigure}{0.19\textwidth}
      \centering
      \includegraphics[width=\linewidth]{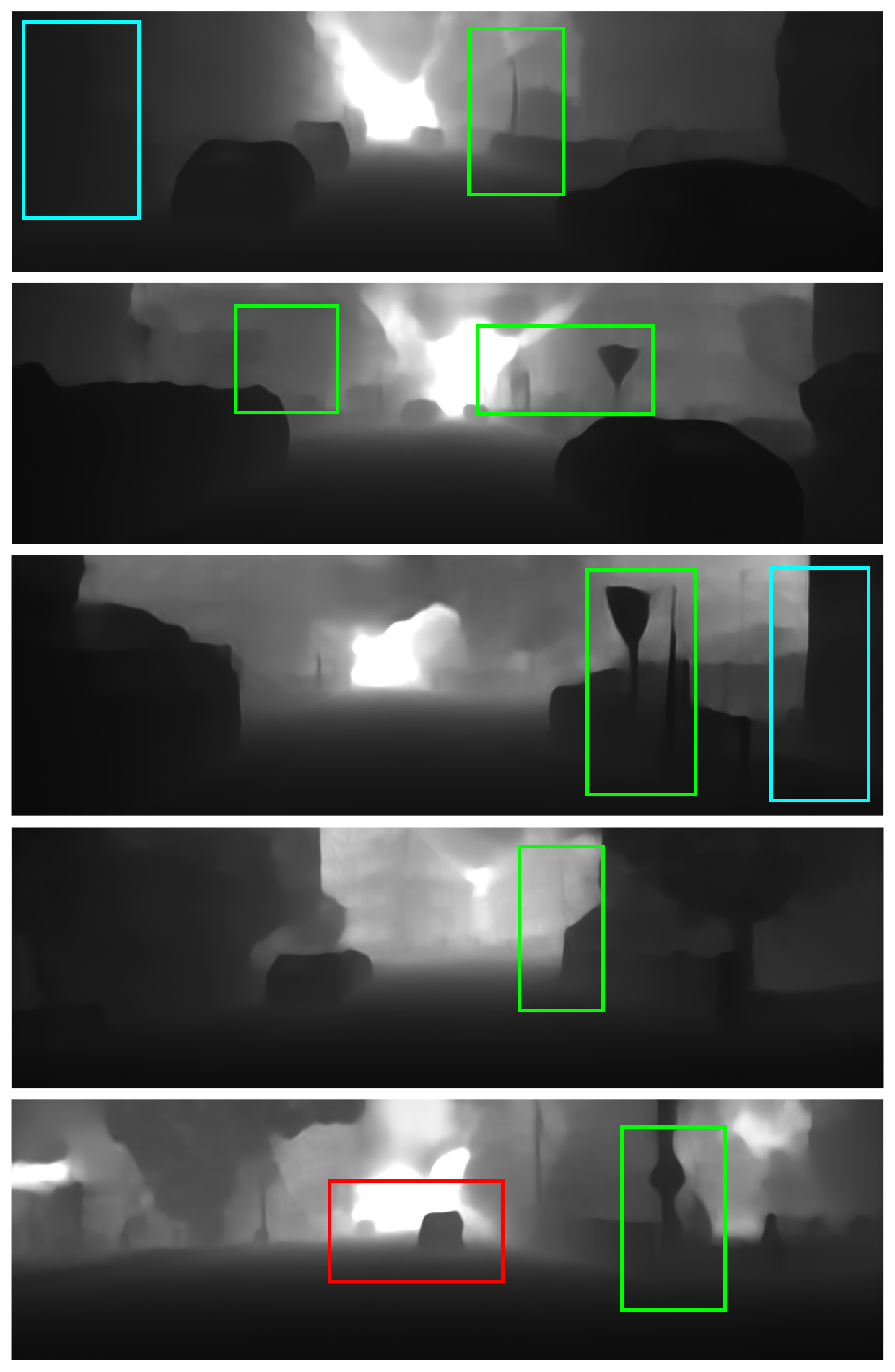}
      \caption{MonoDepth2 \cite{monodepth2}}
    \end{subfigure}
    \begin{subfigure}{0.19\textwidth}
      \centering
      \includegraphics[width=\linewidth]{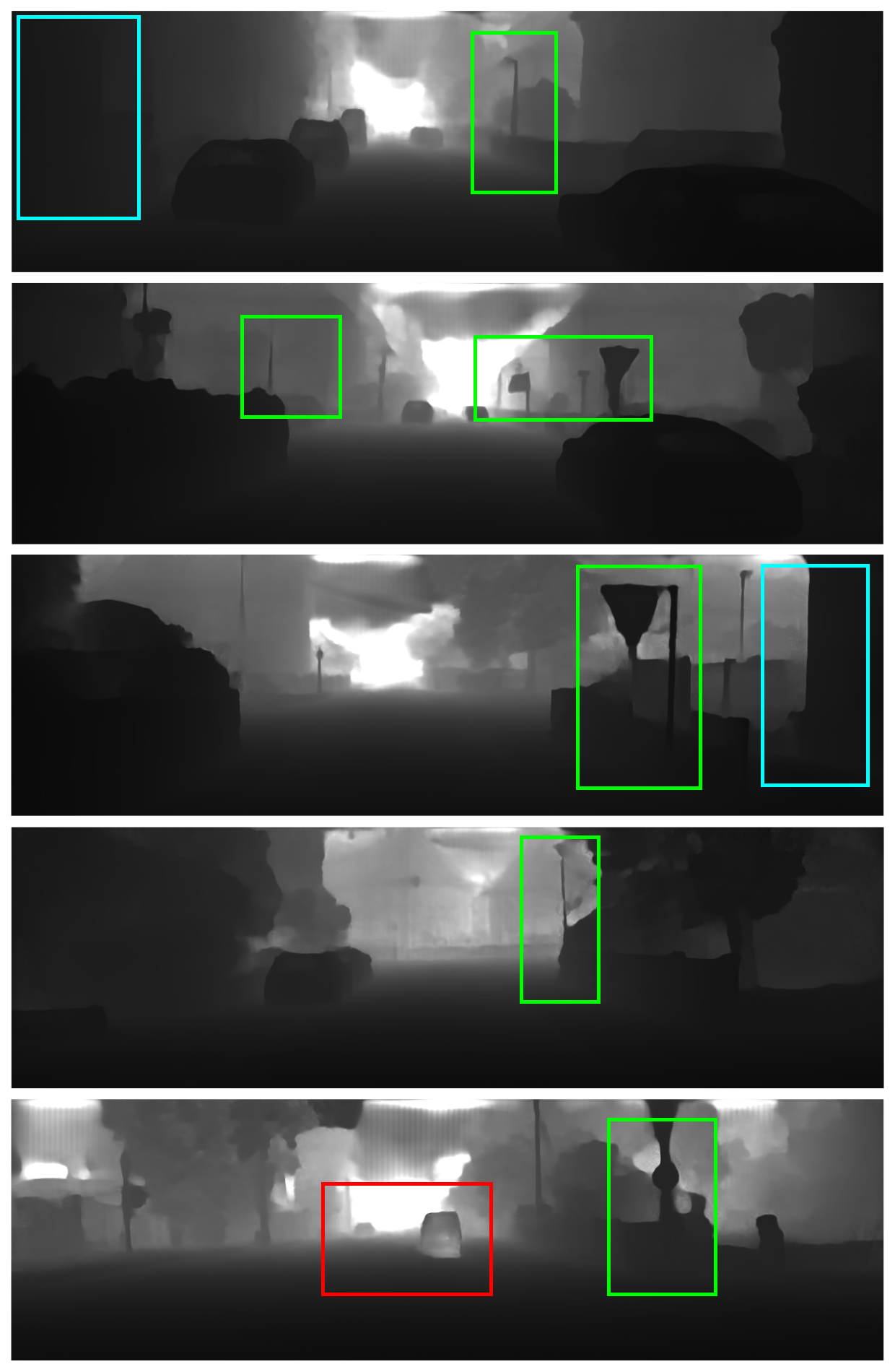}
      \caption{PackNet \cite{packnet}}
    \end{subfigure}
    \begin{subfigure}{0.19\textwidth}
      \centering
      \includegraphics[width=\linewidth]{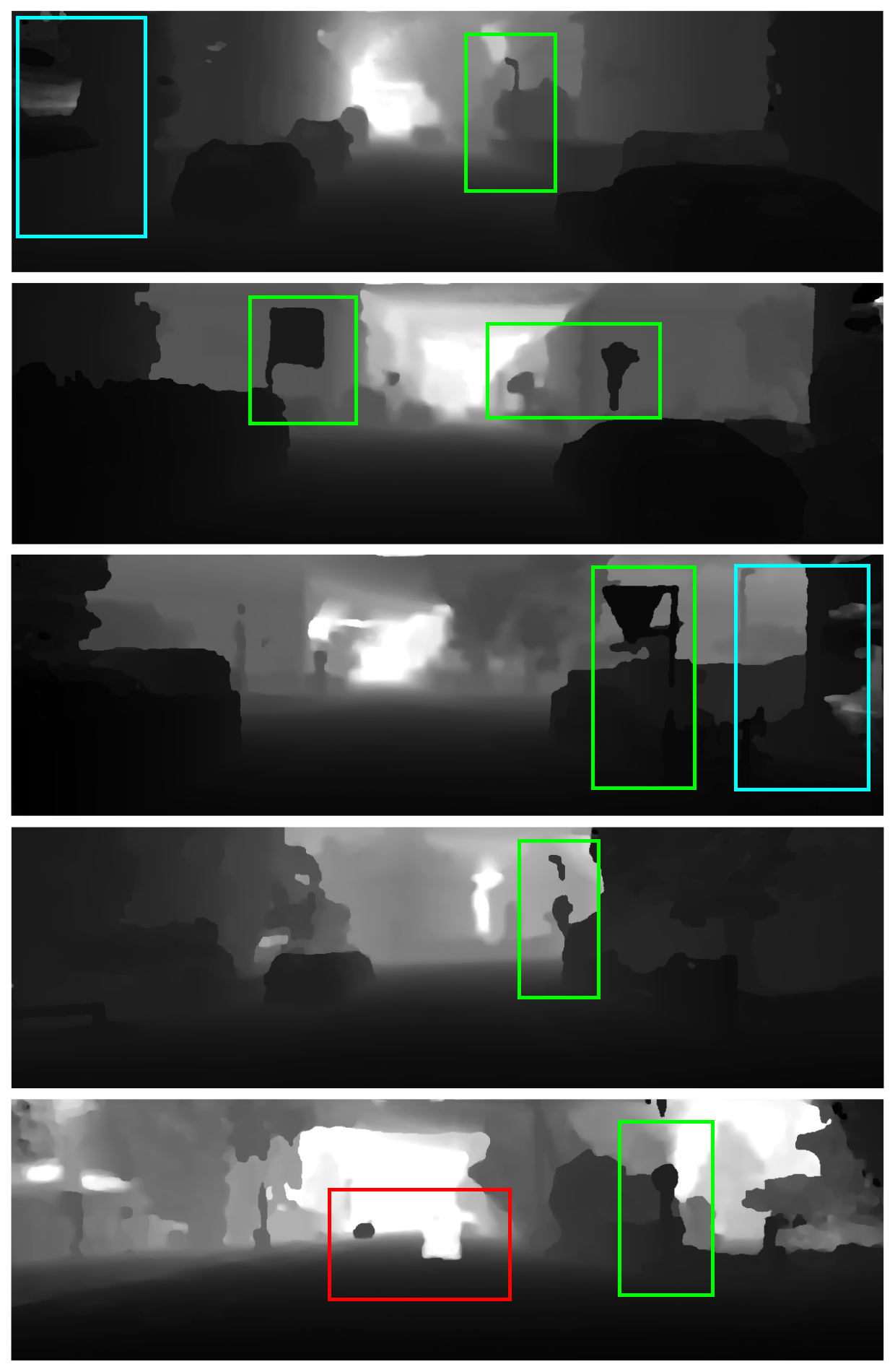}
      \caption{Baseline \cite{acmp}}
    \end{subfigure}
    \begin{subfigure}{0.19\textwidth}
      \centering
      \includegraphics[width=\linewidth]{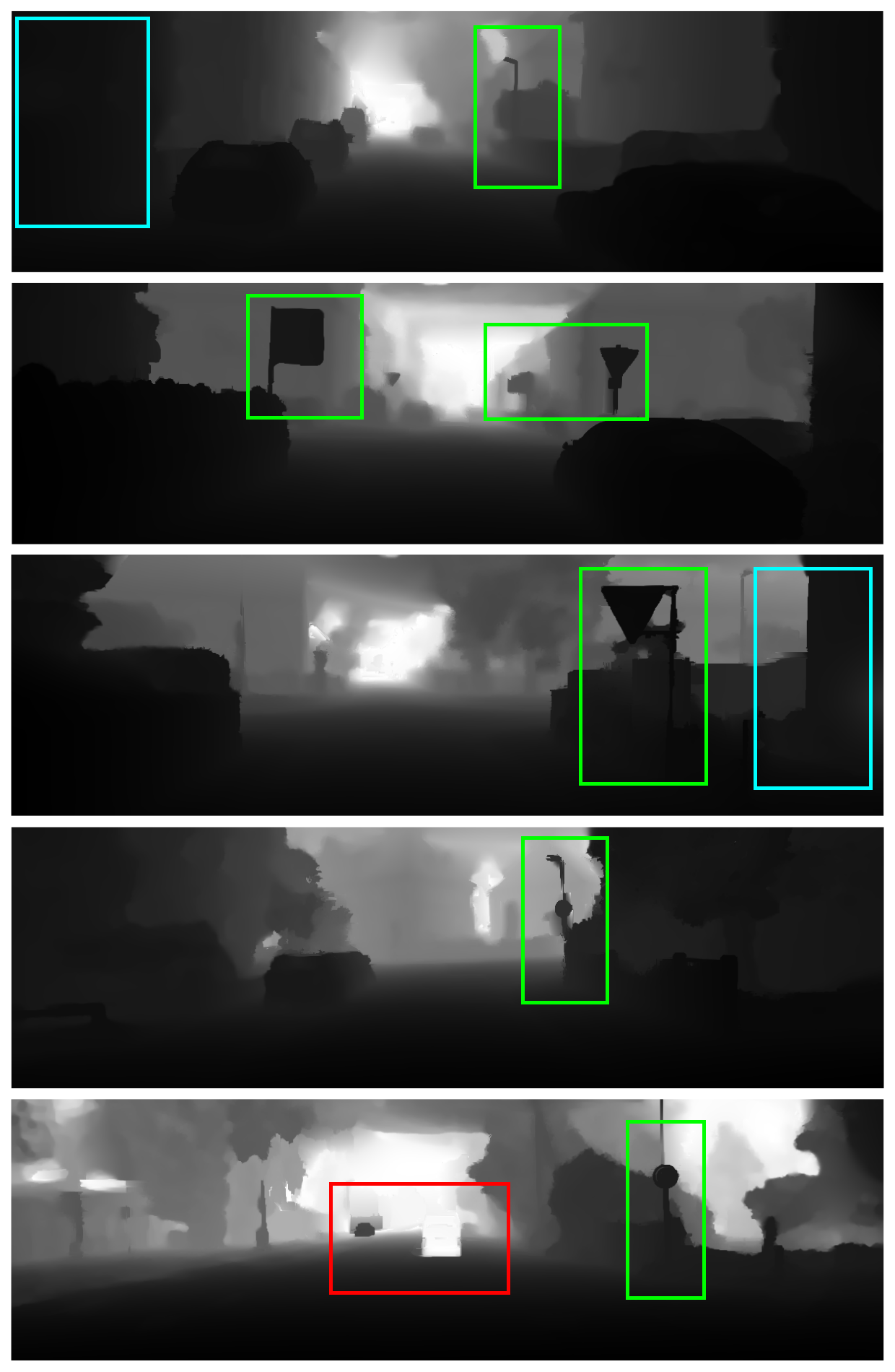}
      \caption{Ours}
    \end{subfigure}
    \caption{Qualitative results on the KITTI dataset. Bounding box legend: \textcolor{green}{Green} - traffic signals detail, \textcolor{cyan}{Cyan} - textureless areas, \textcolor{red}{Red} - moving objects. Differences are best viewed when zoomed in.}
    \label{fig:qual}
\end{figure*}

The proposed framework has been also qualitatively evaluated against state of the art. Fig.~\ref{fig:qual} shows the depth maps computed by two self-supervised monocular networks (\cite{monodepth2, packnet}), the baseline~\cite{acmp} and the presented method. It can be seen that while learning-based approaches provide a smooth and visually pleasant results, they nevertheless lose important details in fine-grained structures such as light poles and traffic signals. Moreover, the local-only optimization strategy of~\cite{acmp} introduces noise and artifacts in large textureless areas. On the other hand, the proposed method can effectively combine these two advantages.

A typical failure case of MVS algorithms is the presence of moving objects. In this situation, the direct image-to-depth mapping learned by monocular networks can still predict a reliable result. A recent work~\cite{monorec} proposed an effective strategy to deal with dynamic objects in MVS as well, which can be applied to obtain accurate depth. However, this issue is mitigated when the final goal is 3D reconstruction, since moving objects are typically masked away with off-the-shelf segmentation networks~\cite{deepvlab3}, as they do not belong to the static environment to be reconstructed (Fig.~\ref{fig:qual}, last row).

\begin{figure*}
    \centering
    \includegraphics[width=\textwidth]{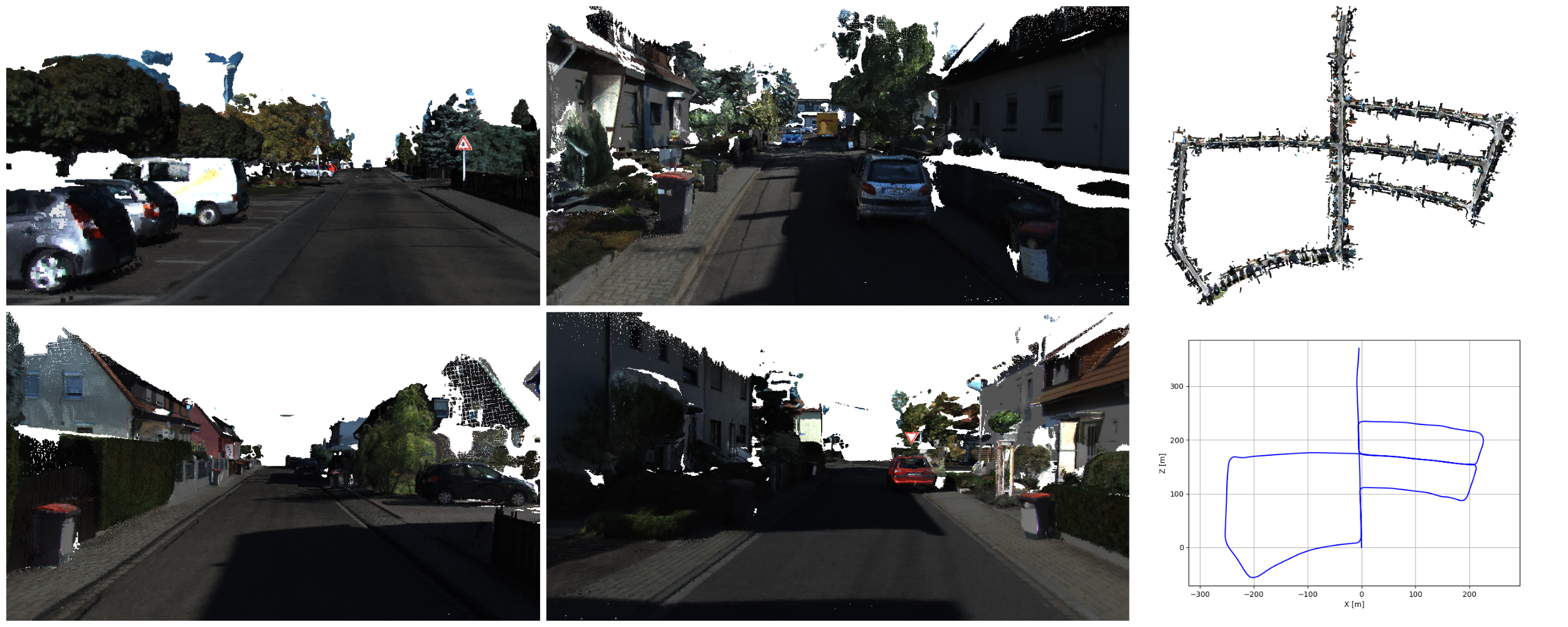}
    \caption{3D reconstruction results from sequence 05 of the KITTI odometry dataset: individual snapshots (left) of the large-scale point cloud (right, top), which is consistent with the estimated camera trajectory (right, bottom).}
    \label{fig:3d}
\end{figure*}

Finally, qualitative 3D reconstruction results on a whole sequence from the KITTI odometry dataset are shown in Fig.~\ref{fig:3d}. Pixelwise depth and normals are computed with the proposed PatchMatch MVS algorithm and back-projected in 3D with the simple procedure of Sec.~\ref{sec:pcl}. The large-scale point cloud (right, top) is consistent with the estimated trajectory (right, bottom), while individual snapshots (left) show that the 3D model is dense and well-textured. The incomplete parts are mainly due to the severely limited field of view of the frontal camera in the dataset.


\section{CONCLUSION}
\label{sec:conc}

In this work, a complete pipeline for monocular 3D reconstruction of urban scenarios is proposed. Three key improvements to a core MVS algorithm are presented: (i) initialization with keypoints from visual SLAM; (ii) a novel geometric consistency loss; (iii) a multi-scale interaction between local PatchMatch optimization and a confidence-based global refinement method. Multi-view consistent depths and normals are then fused into a large-scale point cloud. The novel contributions are evaluated and ablated on the KITTI dataset, showing better performances compared to state of the art MVS methods and monocular depth networks.

{\small
\bibliographystyle{IEEEtran}
\bibliography{egbib}
}

\end{document}